\PassOptionsToPackage{unicode}{hyperref}
\PassOptionsToPackage{hyphens}{url}
\PassOptionsToPackage{dvipsnames,svgnames,x11names}{xcolor}
\documentclass[
  11pt,
]{article}
\usepackage{amsmath,amssymb}
\usepackage{iftex}
\ifPDFTeX
  \usepackage[T1]{fontenc}
  \usepackage[utf8]{inputenc}
  \usepackage{textcomp} % provide euro and other symbols
\else % if luatex or xetex
  \usepackage{unicode-math} % this also loads fontspec
  \defaultfontfeatures{Scale=MatchLowercase}
  \defaultfontfeatures[\rmfamily]{Ligatures=TeX,Scale=1}
\fi
\usepackage{lmodern}
\ifPDFTeX\else
\fi
\IfFileExists{upquote.sty}{\usepackage{upquote}}{}
\IfFileExists{microtype.sty}{% use microtype if available
  \usepackage[]{microtype}
  \UseMicrotypeSet[protrusion]{basicmath} % disable protrusion for tt fonts
}{}
\makeatletter
\@ifundefined{KOMAClassName}{% if non-KOMA class
  \IfFileExists{parskip.sty}{%
    \usepackage{parskip}
  }{% else
    \setlength{\parindent}{0pt}
    \setlength{\parskip}{6pt plus 2pt minus 1pt}}
}{% if KOMA class
  \KOMAoptions{parskip=half}}
\makeatother
\usepackage{xcolor}
\usepackage[margin=1in]{geometry}
\usepackage{listings}

\providecommand{\tightlist}{%
  \setlength{\itemsep}{0pt}\setlength{\parskip}{0pt}}
\usepackage{newunicodechar}
\newunicodechar{−}{\ensuremath{-}}
\newunicodechar{×}{\ensuremath{\times}}
\newunicodechar{±}{\ensuremath{\pm}}
\usepackage[normalem]{ulem}
\usepackage{xcolor}
\usepackage{float}
\usepackage{indentfirst}
\usepackage{graphicx}
\usepackage{booktabs}
\usepackage{caption}
\usepackage{upquote}
\newcommand{\refstart}{\par\setlength{\parindent}{-1.5em}\setlength{\leftskip}{1.5em}}
\newcommand{\refend}{\par\setlength{\parindent}{0pt}\setlength{\leftskip}{0pt}}
\ifLuaTeX
  \usepackage{selnolig}  % disable illegal ligatures
\fi
\IfFileExists{bookmark.sty}{\usepackage{bookmark}}{\usepackage{hyperref}}
\IfFileExists{xurl.sty}{\usepackage{xurl}}{} % add URL line breaks if available
\hypersetup{
  pdftitle={Can an LLM dream? Testing the overfitted brain hypothesis on an agent with text memory},
  pdfauthor={Eric Zou, University of Waterloo},
  colorlinks=true,
  linkcolor={black},
  filecolor={Maroon},
  citecolor={Blue},
  urlcolor={blue},
  pdfcreator={LaTeX via pandoc}}

\title{Can an LLM dream? Testing the overfitted brain hypothesis on an
agent with text memory}
\author{Eric Zou, University of Waterloo}
\date{}

\begin{document}
\maketitle
\begin{abstract}
Hoel (2021) proposed that dreams distort replayed experience so that the
brain does not overfit to recent events. We test this hypothesis on an
LLM agent (Claude Opus 5.5) that stores what it learns in a text memory
and rewrites that memory each night from the day's conversation
transcript. The agent goes through ten simulated days of conversation
built from LongMemEval. We corrupt 5, 10, 20 or 40 percent of the
transcript's turns before each nightly rewrite in one of two ways.
Detail corruption changes one or two details in a turn to plausible
wrong ones, or replaces the turn with one by the same speaker from an
earlier day. Random-text corruption replaces half of the selected turns
with a passage of Moby-Dick and inserts Moby-Dick words into the other
half. A size limit on the memory, rising from 19,500 to 117,000 tokens
over ten nights, set from human free-recall data, forces it to forget.
We learned that corruption did not help the agent apply what it learned
in earlier conversations to new tasks in any of 30 runs (ten conditions,
three different conversation histories). Replacing details in 40 percent
of turns lowered recall from 76 to 60 percent. Random text had no
consistent effect on recall, with a paired difference of 0 ± 2 points at
40 percent, because the model recognized it as corrupted and left it out
of memory. Replaced details were stored as fact even when another turn
held the correct value, and the agent answered 40 percent of those
questions correctly, compared with 75 percent without corruption. These
results suggest that a pretrained model can separate corrupted text from
real events, so a fair test needs a learner that cannot make this
separation as easily, such as a model fine-tuned on the corrupted days
or an agent whose days take place in an invented world.
\end{abstract}

\hypertarget{introduction}{%
\section{1. Introduction}\label{introduction}}

Hoel (2021) proposed the overfitted brain hypothesis. A day of
experience is a small and repetitive sample, so a brain that learns from
it risks overfitting, as a neural network does when trained on a narrow
dataset. Hoel argued that dreams reduce this risk by replaying the day
in a distorted form, which acts like the noise injection and dropout
used to regularize deep networks. Training a network on noisy copies of
its inputs is equivalent to adding a smoothness penalty to the fit,
known as Tikhonov regularization (Bishop, 1995). Figure~1 illustrates
this with a curve fit. The hypothesis predicts that distorted replay
costs little on memorized material and improves performance on new
situations.

LLM agents now have a comparable nightly process. Letta's sleep-time
compute (Lin et al., 2025) lets an agent rewrite its stored context
between sessions, and Behrouz et al.~(2026) argue that language models
need a sleep phase to consolidate what they learn. Most agents store
what they learn as text that the model reads at the start of each
session. We treat a nightly rewrite of that text as a form of learning
and test whether corrupting the transcript it reads from has the effects
Hoel predicts.

\begin{figure}[H]
\centering
\includegraphics[width=0.8\linewidth,height=0.4\textheight,keepaspectratio]{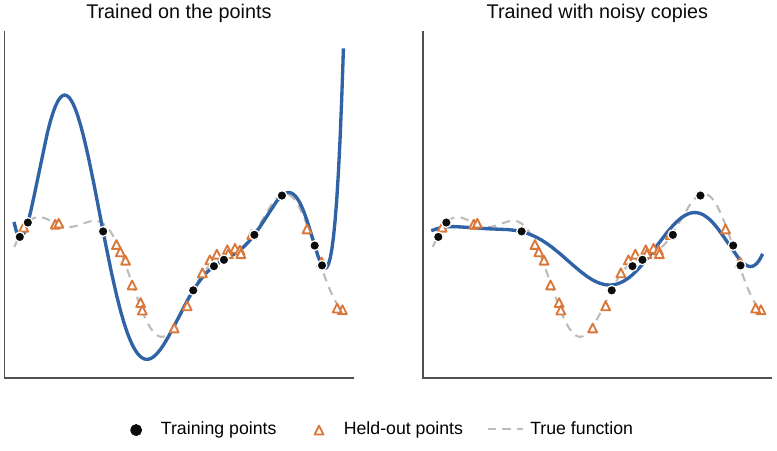}
\caption{An illustration of the overfitted brain hypothesis, not our data. Ten training points (filled circles) come from a smooth function with a small amount of noise. The left panel fits a flexible curve to only these points, and the curve passes through every point and swings between them. The right panel fits the same curve to many noisy copies of the points, and the curve is smoother and closer to the held-out points (open triangles). Our corrupted conditions apply the same idea to the text the model rewrites into memory each night.}
\end{figure}

We expected corruption to help, since noisy input might push the rewrite
toward general lessons and away from copies of the transcript. Instead,
corrupted details lowered recall, and no type or level of corruption
improved transfer to new tasks. Our contributions are:

\begin{itemize}
\tightlist
\item
  An evaluation setup for nightly memory consolidation in LLM agents,
  with a memory size limit set from human recall data and two baselines,
  one with no rewrite and one with an uncorrupted rewrite. The code,
  data and a log of every design decision are public at
  \href{https://github.com/EricJujianZou/llm-dream}{github.com/EricJujianZou/llm-dream}.
\item
  Recall and transfer results for two types of corruption at four rates,
  over 30 runs with Claude Opus 5.5.
\item
  Evidence for why corruption did not help. The model recognizes random
  text as corrupted and drops it, and it stores plausible wrong details
  as fact even when the transcript also contains the correct value.
\end{itemize}

\hypertarget{related-work}{%
\section{2. Related work}\label{related-work}}

\textbf{Dreams as regularization.} Hoel (2021) proposed the overfitted
brain hypothesis and reviewed evidence from sleep deprivation and dream
content. He states the prediction directly: ``According to the OBH
{[}overfitted brain hypothesis{]}, memorization should be less affected
by sleep deprivation than generalization.''

\textbf{Sleep for language models.} Lin et al.~(2025) introduced
sleep-time compute, in which an agent processes its stored context
between user requests so that it has less work to do when a request
arrives. Behrouz et al.~(2026) added a sleep phase to language model
training that includes replay and found that it improved retention.
Neither work compares corrupted replay with uncorrupted replay.

\textbf{Long-term memory benchmarks.} LongMemEval (Wu et al., 2024) is a
benchmark of long chat histories with questions about what the user
said, including questions that need facts from several sessions. We use
its sessions and questions to build the agent's days.

\textbf{False memory and misinformation.} Roediger and McDermott (1995)
showed that people who study a list of related words often recall a
related word that was not on the list (the DRM task). Loftus and Palmer
(1974) showed that a wrong detail given after an event changes what
people later report about it. Cao, Schooler and Zafarani (2025) ran
seven human memory tasks on GPT-4 and two smaller open models. GPT-4
recognized every studied DRM word and labeled 11 percent of the related
words that were never shown as studied, and it showed no primacy or
recency effect when recalling a list. Xie et al.~(2024) found that LLMs
accept coherent conflicting evidence in their context but still favor
evidence that matches their pretraining.

\hypertarget{method}{%
\section{3. Method}\label{method}}

\hypertarget{agent-and-memory}{%
\subsection{3.1 Agent and memory}\label{agent-and-memory}}

The agent is Claude Opus 5.5 (claude-opus-5-5, medium effort), run
through Claude Code in headless mode. Its weights do not change during a
run. Its memory is a list of items, each with a one-line summary and a
set of details, and the model reads the whole memory at the start of
each day. Each night the model reads the day's transcript and the
current memory and returns edits that add, update or merge items. It
also gives each item an importance score from 1 to 5. Appendix~C shows
one memory item.

\hypertarget{simulated-days}{%
\subsection{3.2 Simulated days}\label{simulated-days}}

We built three ten-day conversation histories from LongMemEval sessions.
Bohn and Short (2009) estimate that an American reads or hears about
100,000 words a day, which is about 130,000 tokens, so each simulated
day is 130,000 tokens long. Of these, 126,000 tokens come from 27 to 39
LongMemEval sessions, and 4,000 tokens come from short conversations
written for the transfer test described below. Every condition runs on
the same three histories, which makes comparisons between conditions
paired.

Each history has 100 LongMemEval questions. 84 are answerable, and 22 of
those need facts from more than one session. The other 16 ask about
things the user never said, and the correct response is to decline. We
excluded LongMemEval's temporal-reasoning questions because mixing
sessions from different users breaks their timelines.

The transfer test checks whether the agent can apply something it
learned from earlier conversations to a task it has not seen. Each
history contains 12 facts about the user that change how the assistant
should help them. Each fact comes out in 2 or 3 short conversations
spread over the ten days, and none of these conversations states it as
an instruction. For example, the user in one history only ever has about
15 minutes at a time. They ask how to repot a plant, and in a later
message they say they had to leave it halfway with its roots exposed
because they ran out of time. A model with no memory gives one long
plan, while an agent that learned the fact splits the work into short
parts with clear stopping points. Claude Opus 5.5 wrote the facts and
conversations, and we kept a fact only if the model acted on it in fewer
than 20 percent of attempts with an empty memory. After the last night
the agent gets two tasks per fact, one similar to the conversations
where it came out and one of a different kind. The score is the percent
of facts the agent acts on.

\hypertarget{conditions}{%
\subsection{3.3 Conditions}\label{conditions}}

We ran ten conditions (Figure~2):

\begin{itemize}
\tightlist
\item
  \textbf{No rewrite.} The memory holds the most recent turns of the
  day, cut to the size limit.
\item
  \textbf{Clean rewrite.} The model rewrites its memory from the
  uncorrupted transcript.
\item
  \textbf{Detail corruption at 5, 10, 20 and 40 percent of turns.} Each
  selected turn gets one of three edits before the rewrite. A detail is
  swapped with one from another turn in the same 25,000-token section of
  the transcript, one or two plausible new details are invented, or the
  whole turn is replaced with a turn by the same speaker from an earlier
  day. Night 1 has no earlier day, so turns chosen for replacement on
  night 1 get invented details instead.
\item
  \textbf{Random-text corruption at 5, 10, 20 and 40 percent of turns.}
  Half of the selected turns are replaced by a passage of Moby-Dick
  (Melville, 1851) of the same length. The other half get Moby-Dick
  words inserted at random positions (Table~1), adding 30 percent to
  their length.
\end{itemize}

\begin{figure}[H]
\centering
\includegraphics[width=1\linewidth,height=0.4\textheight,keepaspectratio]{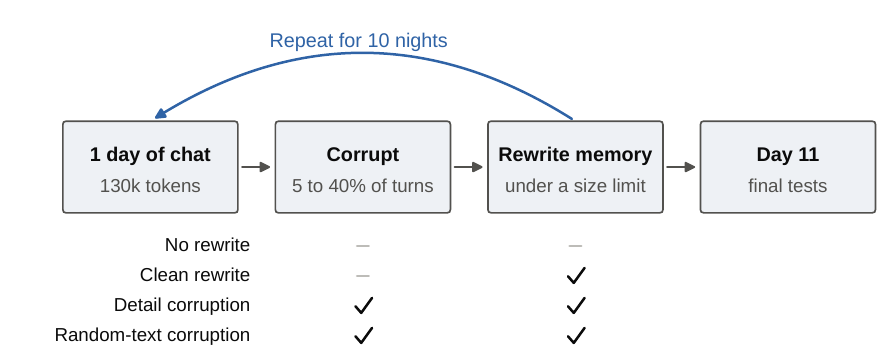}
\caption{Method. Each simulated day adds about 130,000 tokens of conversation. In the corrupted conditions, 5 to 40 percent of the day's turns are corrupted before the model rewrites its memory under a size limit. This repeats for 10 nights, and day 11 holds the final tests. The ticks under the boxes show which steps each condition runs. With no rewrite, memory holds the most recent turns that fit under the limit.}
\end{figure}

A turn is one message from the user or the assistant, and each history
holds about 3,605 turns across the ten days that are rewritten. Summed
over all nights, the four rates corrupt about 181, 361, 720 and 1,430 of
these turns per history. All rewrite conditions use the same prompt
(Appendix~A). Claude Opus 5.5 makes the detail edits from a separate
prompt (Appendix~B).

\begin{table}[H]
\centering\small
\caption{Corruption examples. One real turn from each corrupted condition. Struck-through text was removed and blue text was added.}
\begin{tabular}{p{0.24\linewidth}p{0.70\linewidth}}
\toprule
\textbf{Condition} & \textbf{Turn after corruption} \\
\midrule
Detail corruption & What U.S. state produces the most \sout{peaches} \textcolor{blue}{cherries}? \\
Random-text corruption & \textcolor{blue}{about} What \textcolor{blue}{whale-ship} inspired the lyrics of Windy's ... \\
\bottomrule
\end{tabular}
\end{table}

\hypertarget{memory-size-limit}{%
\subsection{3.4 Memory size limit}\label{memory-size-limit}}

The memory may hold 19,500 tokens after night 1, rising to 117,000
tokens after night 10. The night 1 limit keeps 15 percent of a day, the
middle of the 10 to 25 percent range reported for free recall of a day's
details after one night (Brewer, 1988; Diekelmann and Born, 2010). Later
limits add what remains of each earlier day. We set the share kept by a
day \(t\) nights old by hand, at 15, 12, 10, 9, 8.5, 8, 7.5, 7, 6.5 and
6.5 percent for \(t = 1\) to \(10\). These values were not fitted to
data. They lie close to \(15\,t^{-0.4}\) percent, a power law of the
kind that describes human forgetting (Wixted and Ebbesen, 1991). People
lose verbatim detail before gist (Brainerd and Reyna, 2002), so when the
memory is over the limit our code removes details first. It starts from
the lowest-importance items, and among items of equal importance it
starts from the one least recently written, and deletes whole items only
after every item is down to its summary line.

\hypertarget{measures}{%
\subsection{3.5 Measures}\label{measures}}

Each question is asked on the evening of the day its last supporting
session appears, with that day's transcript available. It is asked again
the next morning with only the memory available, and a third time after
night 10.

\begin{itemize}
\tightlist
\item
  \textbf{Final recall} is the percent of the 84 answerable questions
  answered correctly after night 10.
\item
  \textbf{Multi-session recall} is final recall on the 22 questions that
  need facts from more than one session.
\item
  \textbf{Overnight loss} is evening accuracy minus morning accuracy, in
  points.
\item
  \textbf{Transfer} is the percent of the 12 facts the agent acts on,
  reported separately for similar and different tasks.
\end{itemize}

Day 11 contains three standard memory tests, presented in ordinary
conversation with no announcement: the DRM task, recall of an ordered
list (serial position), and a fan-effect task from Cao, Schooler and
Zafarani (2025) in which each person is paired with one or more places.
The model named the missing words of the classic DRM lists in pilot
runs, so we built new lists from the University of South Florida
free-association norms (Nelson et al., 2004). The tests are given on the
evening of day 11 and again after night 11.

Each answer is graded by gpt-5.6-luna with LongMemEval's grading
prompts, and a second time by Claude Opus 5.5. The two graders agree on
93.6 percent of 8,040 answers.

All 30 runs (ten conditions, three histories) finished. They made 20,323
calls to Claude Opus 5.5, about 1,990 USD at list price, and 15,733
grading calls costing 12 USD.

\hypertarget{results}{%
\section{4. Results}\label{results}}

Values are mean ± SD over the three histories.

\begin{table}[H]
\centering\small
\caption{Results per condition. Recall is in percent and overnight loss is in points. Answer turns hit is the percent of turns containing an answer that the corruption changed.}
\begin{tabular}{lcccc}
\toprule
\textbf{Condition} & \textbf{Final recall} & \textbf{Multi-session recall} & \textbf{Overnight loss} & \textbf{Answer turns hit} \\
\midrule
No rewrite & 14~±~4 & 0~±~0 & 26~±~1 & 0 \\
Clean rewrite & 76~±~5 & 71~±~7 & 8~±~2 & 0 \\
Detail 5\% & 76~±~5 & 67~±~9 & 8~±~2 & 6~±~2 \\
Detail 10\% & 71~±~2 & 61~±~5 & 13~±~1 & 10~±~1 \\
Detail 20\% & 69~±~2 & 59~±~5 & 16~±~4 & 22~±~2 \\
Detail 40\% & 60~±~4 & 45~±~5 & 21~±~2 & 37~±~3 \\
Random text 5\% & 78~±~4 & 76~±~3 & 10~±~5 & 5~±~3 \\
Random text 10\% & 77~±~5 & 74~±~13 & 10~±~2 & 14~±~2 \\
Random text 20\% & 70~±~5 & 61~±~17 & 14~±~1 & 20~±~5 \\
Random text 40\% & 76~±~3 & 73~±~12 & 14~±~3 & 40~±~5 \\
\bottomrule
\end{tabular}
\end{table}

\hypertarget{recall}{%
\subsection{4.1 Recall}\label{recall}}

Final recall falls to 14~±~4 percent without the rewrite. The size limit
therefore removes most of what the agent saw. The clean rewrite reaches
76~±~5. Detail corruption lowers final recall as the rate rises, to
60~±~4 at 40 percent, a paired difference of −17~±~5 points from the
clean rewrite (Figure~3). The difference at 20 percent (−7~±~7) is
within run-to-run variation. Random text has no consistent effect, and
its paired difference from the clean rewrite at 40 percent is 0~±~2
points.

\begin{figure}[H]
\centering
\includegraphics[width=1\linewidth,height=0.4\textheight,keepaspectratio]{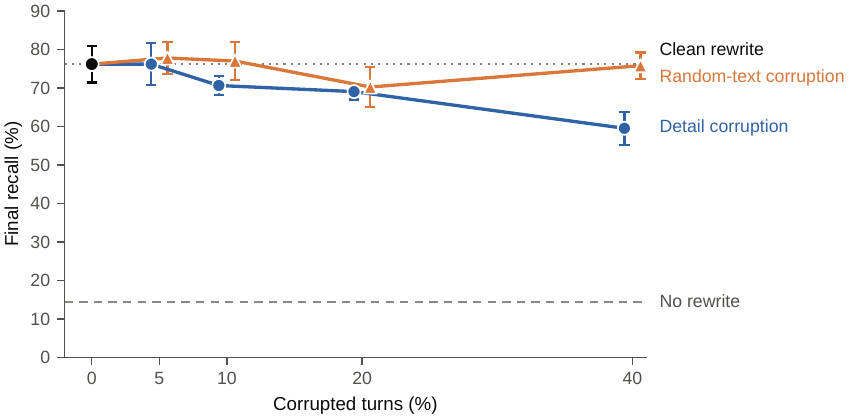}
\caption{Final recall against the percent of corrupted turns for detail and random-text corruption. Both lines start at the clean rewrite. The dotted line marks the clean rewrite and the dashed line marks the no-rewrite baseline. Bars show one standard deviation over the three histories. Differences of 2 points or less are smaller than the standard deviation across the three histories, which is 5 points for the clean rewrite.}
\end{figure}

The flat random-text result is not explained by random text missing the
answers. At 40 percent it changed 40~±~5 percent of the turns that
contain an answer, compared with 37~±~3 percent for detail corruption.

\hypertarget{why-the-two-types-of-corruption-differ}{%
\subsection{4.2 Why the two types of corruption
differ}\label{why-the-two-types-of-corruption-differ}}

The memory in the 40 percent random-text condition held 112 notes
marking content as corrupted after night 9 of one history, and no text
from Moby-Dick. The memory in the 40 percent detail condition held one
such note. These counts come from one run per condition. The rewrite
prompt (Appendix~A) never mentions corruption, noise or unrelated text,
so the model added these notes without being told that corruption could
occur.

Overnight loss rises from 8~±~2 points with the clean rewrite to 21~±~2
points with 40 percent detail corruption, and to 14~±~3 points with 40
percent random text (Figure~D1). Detail corruption therefore does most
of its damage during the rewrite.

Many answers appear in more than one turn. We grouped questions by what
happened to these turns, pooling the three histories. When random text
replaced every turn containing the answer (23 of 252 questions), the
agent answered 8 correctly, compared with 12 for the same questions
under the clean rewrite. When some turns with the answer were corrupted
and at least one was left intact, the agent answered 79 percent of 57
questions correctly with 40 percent random text and 40 percent of 67
questions correctly with 40 percent detail corruption, compared with 75
percent under the clean rewrite. The agent recovered the answer from an
intact turn when the other copies were random text, but a plausible
wrong detail in another copy often overrode the intact turn.

Random text cost little recall, as Hoel predicts for distorted replay,
but detail corruption cost 17 points at 40 percent.

\hypertarget{multi-session-questions-and-transfer}{%
\subsection{4.3 Multi-session questions and
transfer}\label{multi-session-questions-and-transfer}}

Multi-session recall is 71~±~7 with the clean rewrite and 0 in every
history with no rewrite (Figure~D2). Detail corruption lowers it to 67,
61, 59 and 45 percent at the four rates. Random text gives 76, 74, 61
and 73 percent, with SDs up to 17 points, too wide to separate any
random-text condition from the clean rewrite.

\begin{figure}[H]
\centering
\includegraphics[width=1\linewidth,height=0.31\textheight,keepaspectratio]{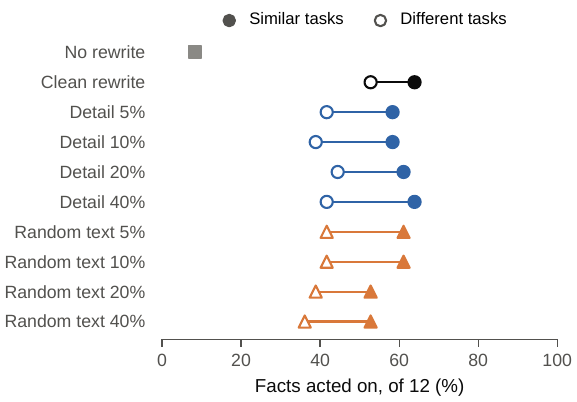}
\caption{Percent of the 12 facts about the user acted on per condition, on similar tasks (filled) and different tasks (hollow). The grey square is No rewrite, where both values are 8 percent. Values are means over the three histories.}
\end{figure}

The clean rewrite acts on 64~±~21 percent of facts on similar tasks and
53~±~17 percent on different tasks (Figure~4). The corrupted conditions
act on 53 to 64 percent on similar tasks and 36 to 44 percent on
different tasks. No rewrite acts on 8~±~8 and 8~±~14 percent. Corruption
did not make the memory more general. The gap between similar and
different tasks is 11~±~13 points with the clean rewrite and 14 to 22
points with corruption, so corruption did not reduce it.

The results do not support Hoel's prediction that distorted replay
improves performance on new situations. We claim only that no corrupted
condition outperformed the clean rewrite beyond run-to-run variation,
because the variation across histories is wide.

\hypertarget{other-results}{%
\subsection{4.4 Other results}\label{other-results}}

Recall under the clean rewrite stays between 59 and 89 percent for
questions 1 to 10 nights old, with no downward trend (Figure~D3). All 30
runs ended within 1,200 tokens of the size limit. The limit never
deleted a whole item in any run and only removed details, for example
52~±~26 details over ten nights with the clean rewrite.

The agent never labeled the missing DRM word as studied in any
condition. In Roediger and McDermott (1995), people called the missing
word studied about as often as the real list words, at 0.98 to 1.11
times the rate across conditions. GPT-4 recognized every real list word
and called the missing word studied 11 percent of the time (Cao,
Schooler and Zafarani, 2025). In every run where they could be scored,
the agent recalled 93 to 100 percent of the ordered list and answered 96
to 100 percent of the fan-effect questions correctly, so neither test
can show its effect. The missed list items all came from detail
corruption at 20 and 40 percent. In two runs (detail corruption and
random text, both at 20 percent) the corruption changed the turn
containing the ordered list, and the agent said it had no record of the
list. We excluded these two runs from the serial-position results and
did not count them as forgetting. The fan-effect test has no score in
two random-text runs, one at 20 percent and one at 40 percent. The agent
refused no question in any condition.

\hypertarget{discussion}{%
\section{5. Discussion}\label{discussion}}

\textbf{Why corruption did not help.} Recall was not trivial, because
the questions ask about personal details of LongMemEval users that no
pretraining data contains. The model could still tell a real
conversation from corrupted text. In the one 40 percent random-text run
we counted, its memory held 112 notes marking content as corrupted and
no text from Moby-Dick. Hoel's mechanism assumes a learner whose only
training data is its own experience, so that a dream is the only
alternative version of the world it sees. A pretrained model has read
enough text to recognize a paragraph of Moby-Dick as out of place.
Transfer likely came from pretraining as well. A plausible account is
that the agent applied a fact about the user by reading its notes when
it answered, using general ability it already had, but we did not test
this.

\textbf{Why wrong details cost more than random text.} The rewrite
stored plausible wrong details as fact, and the agent treats its memory
as fact when it answers. This resembles the misinformation effect in
people (Loftus and Palmer, 1974), where a wrong detail given after an
event becomes part of the reported memory. The agent showed no DRM false
memory, because a word that was never written to its memory cannot be
read back from it.

\textbf{Where the hypothesis could still hold.} A fair test needs a
learner whose knowledge comes mainly from its days. A small model
trained from scratch on the days would meet this condition. Our next
experiment takes a different route and fine-tunes a small open model
each night on the clean or corrupted transcript. Days set in an invented
world whose facts contradict common knowledge would also make corrupted
text harder to separate from real events. Wilhelm et al.~(2013) found
that children gained more explicit knowledge from sleep than adults
after learning a motor sequence, and Wilhelm et al.~(2008) found that
sleep improved recall of word pairs about equally in children and
adults. We speculate that a learner with less prior knowledge gains more
from distorted replay, because it cannot dismiss a strange dream as
noise. We found no study that tests this in people.

\hypertarget{limitations}{%
\section{6. Limitations}\label{limitations}}

\begin{itemize}
\tightlist
\item
  We tested only Claude Opus 5.5. A model that does not recognize random
  text as corrupted might store it, and random text would then lower
  recall as detail corruption did.
\item
  The size limit follows one forgetting curve and never deleted a whole
  item. A tighter limit might separate the conditions on forgetting
  instead of on stored errors.
\item
  Three histories give wide variation, and we treat differences within
  it as no result. This includes the 20 percent detail condition on
  final recall, multi-session recall and transfer to different tasks.
\item
  Recall rests on 84 questions per history and multi-session recall on
  22, so one question changes multi-session recall by 4.5 points.
\item
  The transfer test has 12 facts per history, with SDs of 5 to 27
  points. A benefit smaller than that would not show.
\item
  Accuracy is not comparable with published LongMemEval results, because
  our days mix several users' histories, exclude temporal-reasoning
  questions and use a different grader.
\item
  The count of corrupted answer turns includes only edits whose before
  and after text was recorded, so it may undercount detail corruption.
\item
  The counts of notes marking corrupted content come from one run per
  condition.
\item
  Our check against a published Letta sleep-time compute result returned
  a gradable answer for only 53 of 100 problems without sleep-time
  compute and 59 of 100 with it, because the other requests timed out or
  failed with server errors. Accuracy on those problems was 0.19 and
  0.31, against 0.05 and 0.36 read from the published figure, so our
  setup has no confirmed match to a published number.
\item
  Claude Code gave the model the real date. This affects every condition
  equally, and the sessions carry no dates.
\item
  The study tests the hypothesis only for a text memory with fixed
  weights.
\end{itemize}

\hypertarget{acknowledgments}{%
\section{Acknowledgments}\label{acknowledgments}}

The author designed and ran the experiments and drew the conclusions.
Claude (Anthropic), a language model, helped write the experiment code
and draft the text.

\hypertarget{references}{%
\section{References}\label{references}}

\refstart

Behrouz, A., Hashemi, F., Javanmard, A., and Mirrokni, V. (2026).
Language models need sleep: Learning to self-modify and consolidate
memories. \href{https://arxiv.org/abs/2606.03979}{arXiv:2606.03979}

Bishop, C. M. (1995). Training with noise is equivalent to Tikhonov
regularization. \emph{Neural Computation, 7}(1), 108-116.
\url{https://doi.org/10.1162/neco.1995.7.1.108}

Bohn, R. E., and Short, J. E. (2009). \emph{How much information? 2009
report on American consumers}. Global Information Industry Center,
University of California, San Diego.

Brainerd, C. J., and Reyna, V. F. (2002). Fuzzy-trace theory and false
memory. \emph{Current Directions in Psychological Science, 11}(5),
164-169. \url{https://doi.org/10.1111/1467-8721.00192}

Brewer, W. F. (1988). Memory for randomly sampled autobiographical
events. In U. Neisser and E. Winograd (Eds.), \emph{Remembering
reconsidered: Ecological and traditional approaches to the study of
memory} (pp.~21-90). Cambridge University Press.

Cao, Z., Schooler, L., and Zafarani, R. (2025). Analyzing memory effects
in large language models through the lens of cognitive psychology.
\href{https://arxiv.org/abs/2509.17138}{arXiv:2509.17138}

Diekelmann, S., and Born, J. (2010). The memory function of sleep.
\emph{Nature Reviews Neuroscience, 11}(2), 114-126.
\url{https://doi.org/10.1038/nrn2762}

Hoel, E. (2021). The overfitted brain: Dreams evolved to assist
generalization. \emph{Patterns, 2}(5), 100244.
\url{https://doi.org/10.1016/j.patter.2021.100244}

Lin, K., Snell, C., Wang, Y., Packer, C., Wooders, S., Stoica, I., and
Gonzalez, J. E. (2025). Sleep-time compute: Beyond inference scaling at
test-time. \href{https://arxiv.org/abs/2504.13171}{arXiv:2504.13171}

Loftus, E. F., and Palmer, J. C. (1974). Reconstruction of automobile
destruction: An example of the interaction between language and memory.
\emph{Journal of Verbal Learning and Verbal Behavior, 13}(5), 585-589.
\url{https://doi.org/10.1016/S0022-5371(74)80011-3}

Melville, H. (1851). \emph{Moby-Dick; or, The Whale}. Project Gutenberg
eBook 2701.

Nelson, D. L., McEvoy, C. L., and Schreiber, T. A. (2004). The
University of South Florida free association, rhyme, and word fragment
norms. \emph{Behavior Research Methods, Instruments, and Computers,
36}(3), 402-407. \url{https://doi.org/10.3758/BF03195588}

Roediger, H. L., III, and McDermott, K. B. (1995). Creating false
memories: Remembering words not presented in lists. \emph{Journal of
Experimental Psychology: Learning, Memory, and Cognition, 21}(4),
803-814. \url{https://doi.org/10.1037/0278-7393.21.4.803}

Wilhelm, I., Diekelmann, S., and Born, J. (2008). Sleep in children
improves memory performance on declarative but not procedural tasks.
\emph{Learning \& Memory, 15}(5), 373-377.
\url{https://doi.org/10.1101/lm.803708}

Wilhelm, I., Rose, M., Imhof, K. I., Rasch, B., Büchel, C., and Born, J.
(2013). The sleeping child outplays the adult's capacity to convert
implicit into explicit knowledge. \emph{Nature Neuroscience, 16}(4),
391-393. \url{https://doi.org/10.1038/nn.3343}

Wixted, J. T., and Ebbesen, E. B. (1991). On the form of forgetting.
\emph{Psychological Science, 2}(6), 409-415.
\url{https://doi.org/10.1111/j.1467-9280.1991.tb00175.x}

Wu, D., Wang, H., Yu, W., Zhang, Y., Chang, K.-W., and Yu, D. (2024).
LongMemEval: Benchmarking chat assistants on long-term interactive
memory. \emph{International Conference on Learning Representations}
(ICLR 2025). \href{https://arxiv.org/abs/2410.10813}{arXiv:2410.10813}

Xie, J., Zhang, K., Chen, J., Lou, R., and Su, Y. (2024). Adaptive
chameleon or stubborn sloth: Revealing the behavior of large language
models in knowledge conflicts. \emph{International Conference on
Learning Representations} (ICLR 2024).
\href{https://arxiv.org/abs/2305.13300}{arXiv:2305.13300}

\refend

\appendix
\clearpage

\hypertarget{appendix-a.-nightly-rewrite-prompt}{%
\section{Appendix A. Nightly rewrite
prompt}\label{appendix-a.-nightly-rewrite-prompt}}

The system prompt for every rewrite condition, verbatim. The values in
braces are filled in each night.

\begin{lstlisting}
You are the sleeping mind of a personal assistant. Tonight you turn today's experiences into long-term memory. Tomorrow the assistant wakes up with only what you write here. Today's transcript will be gone.

Consolidate the way sleep does in people:
- Keep what came up more than once, today or on earlier days.
- Keep what was emotional, surprising or new.
- Connect today's events to what is already in memory, and merge them into it.
- When something changed, keep the newer version and note that it changed.
- Keep the gist. Exact wording can go.
- Let one-off details that never came back fade.

The memory may hold up to {cap} tokens after tonight. It holds {before} tokens now. After tonight it should hold close to {cap} tokens, so it has to grow by about {room_words} words. Rewriting an existing item replaces its text and adds nothing, so that growth has to come from new items and from details added to the items you rewrite. Fill it with specifics from today's transcript (names, numbers, places, what was said and how it went), not filler.

The memory is a list of items. Each item has an id, a one-line gist, its details, and an importance score from 1 to 5. You will see the existing items with their ids. Reply with your changes only, as JSON:
- "add": new items from today, each with "gist", "details" and "score".
- "update": existing items to rewrite, each with "id", "gist", "details" and "score". Use this when something changed or came up again.
- "merge": groups of existing items that belong together, each with "ids", "gist", "details" and "score". The group becomes one item.
Items you do not mention stay as they are. When the memory is over its limit, details are forgotten from the lowest-scored items first, and whole items are forgotten only once every item is down to its gist, so the score is how you choose what fades.
\end{lstlisting}

\hypertarget{appendix-b.-detail-corruption-prompt}{%
\section{Appendix~B. Detail corruption
prompt}\label{appendix-b.-detail-corruption-prompt}}

The system prompt for the detail edits, verbatim. Our code chooses the
turns and the edit type for each turn, and makes the third edit type
(replacing a turn with one from an earlier day) without the model.

\begin{lstlisting}
You edit chat transcripts for a memory experiment. You will receive numbered turns from today's transcript and a list of edits to make. For each listed turn, rewrite it so it is still fluent and plausible in context but factually different:
- "swap": replace one concrete detail (a name, number, place, date, item, or outcome) with a concrete detail taken from a different turn in the transcript.
- "change": replace one or two concrete details with new, plausible but different ones.
Keep each rewritten turn about the same length and the same speaker. Do not mark the edits. Return JSON only, in the form {"edits": [{"index": 12, "text": "rewritten turn"}, ...]}.
\end{lstlisting}

\hypertarget{appendix-c.-example-memory-item}{%
\section{Appendix~C. Example memory
item}\label{appendix-c.-example-memory-item}}

One item from the clean rewrite after night 1 of the first history. The
full code for the memory is in the repository.

\begin{lstlisting}
{
 "id": "m6",
 "gist": "Sister's wedding on the 28th; user promised a photo slideshow and is behind",
 "details": "Session 6: the sister's wedding was three weeks out when first asked. The user promised a photo slideshow and had nothing. Later, on the 14th, they had about 20 photos and had stalled because the week-based plan had no fixed start date. Original plan: collect 80-120 images (text family for favorites, pull from phone and social media); arrange roughly chronologically; 3-4 songs; build in Google Slides, Canva or iMovie at 5-6 s per photo; review, cut weak shots, sync to music, export a USB backup. Follow-up idea: if it comes up, ask how it went and offer a dated day-by-day checklist. The wedding may be what Noah missed school for.",
 "score": 4
}
\end{lstlisting}

\hypertarget{appendix-d.-additional-figures}{%
\section{Appendix~D. Additional
figures}\label{appendix-d.-additional-figures}}

Figures~D1 to D3 show overnight loss, multi-session recall, and final
recall by the number of nights since the supporting session, for every
condition.

\setcounter{figure}{0}\renewcommand{\thefigure}{D\arabic{figure}}

\begin{figure}[H]
\centering
\includegraphics[width=1\linewidth,height=0.4\textheight,keepaspectratio]{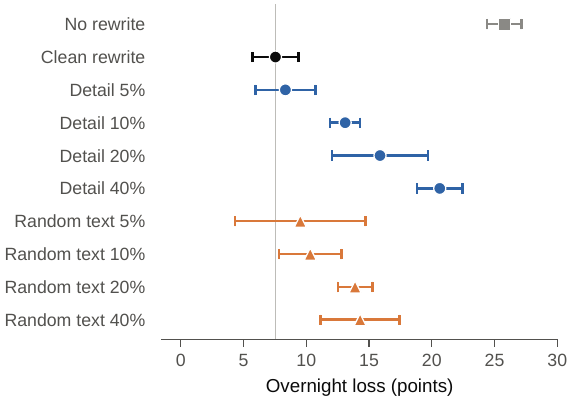}
\caption{Overnight loss per condition. Each point is the mean over the three histories, and the bar spans one standard deviation either side. The vertical line marks the mean of the clean rewrite.}
\end{figure}

\begin{figure}[H]
\centering
\includegraphics[width=1\linewidth,height=0.4\textheight,keepaspectratio]{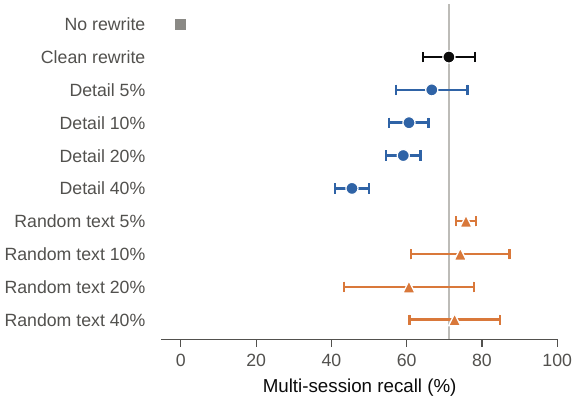}
\caption{Multi-session recall per condition. Each point is the mean over the three histories, and the bar spans one standard deviation either side. The vertical line marks the mean of the clean rewrite.}
\end{figure}

\begin{figure}[H]
\centering
\includegraphics[width=1\linewidth,height=0.4\textheight,keepaspectratio]{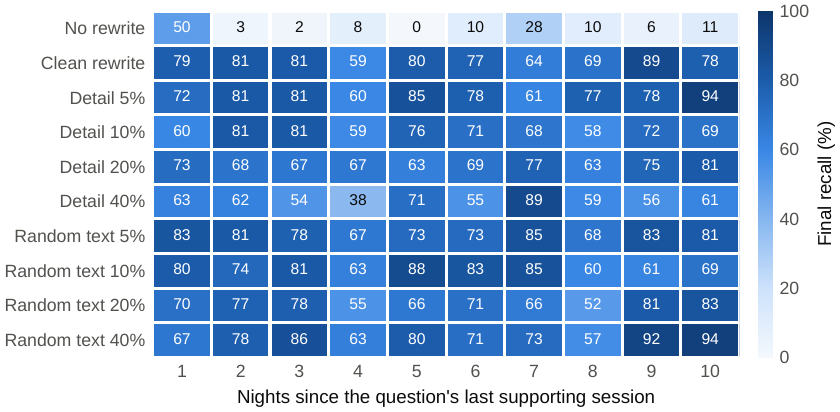}
\caption{Final recall per condition by the number of nights since the question's last supporting session. Values are means over the three histories.}
\end{figure}

\end{document}